\documentclass[runningheads]{llncs}

\usepackage[T1]{fontenc}
\usepackage{graphicx}
\usepackage{amsmath}
\usepackage{mathtools}
\usepackage{graphicx}
\usepackage[colorlinks=true, allcolors=blue]{hyperref}
\usepackage{algorithm} 
\usepackage{algpseudocode}
\usepackage{enumitem}
\usepackage{amsfonts}
\usepackage{amssymb}
\usepackage{booktabs}
\usepackage{verbatim}
\usepackage{multirow}
\usepackage{soul}

\newcommand{\beq}{\begin{equation}}
\newcommand{\eeq}{\end{equation}}
\newcommand{\md}{\mathcal{D}}
\newcommand{\mm}{\mathcal{M}}
\newcommand{\mw}{\mathcal{W}}

\newcommand{\ml}{\mathcal{L}}
\newcommand{\yh}{\hat{y}}

\newcommand{\lgbms}{$\text{LGBM}^{S}$}
\newcommand{\lgbmm}{$\text{LGBM}^{M}$}
\newcommand{\xgbs}{$\text{XGBoost}^{S}$}
\newcommand{\xgbm}{$\text{XGBoost}^{M}$}

\allowdisplaybreaks

\begin{document}
\title{Model Optimization in Imbalanced Regression}


\author{Aníbal Silva\inst{1} \and Rita P. Ribeiro\inst{1,2} \and
Nuno Moniz\inst{2,1}}
\authorrunning{A. Silva et al.}
%
\institute{Faculty of Sciences - University of Porto, Porto, Portugal\\
\email{up201008538@up.pt, rpribeiro@fc.up.pt}
\and 
INESC TEC, Porto, Portugal \\ \email{nmmoniz@inesctec.pt} 
}

\maketitle

\begin{abstract}
Imbalanced domain learning aims 
to produce accurate models in predicting instances that, though underrepresented, are of utmost importance for the domain.
Research in this field has been mainly focused on classification tasks. Comparatively, the number of studies carried out in the context of regression tasks is negligible. 
One of the main reasons for this is the lack of loss functions capable of focusing on minimizing the errors of extreme (rare) values. Recently, an evaluation metric was introduced: Squared Error Relevance Area ($SERA$). This metric posits a bigger emphasis on the errors committed at extreme values while also accounting for the performance in the overall target variable domain, thus preventing severe bias. 
However, its effectiveness as an optimization metric is unknown.
In this paper, our goal is to study the impacts of using $SERA$ as an 
optimization criterion in imbalanced regression tasks. 
Using gradient boosting algorithms as proof of concept,  we perform an experimental study with 36 data sets of different domains and sizes. Results show that models that used $SERA$ as an objective function are practically better than the models produced by their respective standard boosting algorithms at the prediction of extreme values.  This confirms that $SERA$ can be embedded as a loss function into optimization-based learning algorithms 
for imbalanced regression scenarios.

\keywords{Imbalanced Regression . Asymmetric Loss Functions . Model Optimization \and Boosting .}
\end{abstract}

\section{Introduction} \label{sec:int}

Supervised learning assumes there is an unknown function mapping a set of independent to one or more dependent variables. Learning algorithms aim to approximate such an unknown function through optimization processes. A key decision rests on choosing which preference criterion, e.g. a loss function, should be used. Such a decision entails critical definitions and assumptions on what should be considered a successful approximation. Most importantly, we should stress that, these decisions are commonly aimed at minimizing the overall error across the entire domain of the target (dependent) variable of a given data set. By assuming that all values are equally important, traditional optimization processes tend to produce models that have a particular focus on the most common values of the target variable. This is not the goal of many real-world applications that configure imbalanced domain learning tasks. 

In imbalanced learning, the following holds: \textit{i)} the target variable has a non-uniform or skewed distribution; \textit{ii)} the values across the domain of the target variable are not equally important; and \textit{iii)} the focus is on the rare cases, i.e. values that are poorly represented in the data set.
Examples of this type of predictive task spread from classification to regression. They include multiple real-world applications in different areas, such as finance, where the user might be interested in fraud detection, and environmental sciences, to mitigate the occurrence of natural catastrophes, such as floods and hurricanes. 


Focusing on Imbalanced Regression, several challenges impose the non-triviality of predicting extreme values. From a supervised learning perspective, these include two main ones: 
1) the definition of suitable and non-uniform preferences over a continuous and possibly infinite domain of the target variable;
2) map such preference regarding the extreme values into an evaluation metric that would adequately allow model selection and, possibly, optimization.
Regarding the first challenge, a proposal~\cite{ribeiro,imbregex} exists that suggests a mapping of the target variable domain into a well-defined space (the relevance space), which gives information about the relevance of a given instance based on its target value. As for the second challenge, while there are some proposals for specially tailored evaluation metrics ~\cite{linex,linlin,clive} in an imbalanced regression scenario, only a very few works exist on including such metrics in the optimization process.
We focus on this second challenge. In particular, we build on recent work that introduced the Squared Error Relevance Area ($SERA$)~\cite{imbregex} metric. This metric allows for errors of equal magnitude to have different impacts depending on the relevance of the target values. Moreover, while it focuses on errors in cases with extreme target values, it also accounts for the errors committed across all the rest of the target values, preventing a severe bias towards the extreme values. However, despite its demonstrated interest in model selection tasks, it is unclear if it is possible to use it directly in optimization processes.

In this work, our main contribution is to show that $SERA$ can be used as an optimization loss function in machine learning algorithms, with the ability to generalize its predictive power for both average and extreme target value instances. Our demonstration efforts consist of empirical evaluation using gradient boosting algorithms and a test bed of 36 data sets. Results show us that, overall, $SERA$ can be used as an optimization loss function. In addition, when these models optimized with $SERA$ under-perform w.r.t. models optimized via standard loss functions (e.g. $MSE$), the former still have the ability to outperform on extreme values, opening horizons to a broader set of applications in the realm of Imbalance Regression (e.g. Deep Learning).

The paper is organized as follows. In Section~\ref{sec:recentwork}, we provide a review of recent related work regarding imbalanced domain learning. In Section~\ref{sec:imbreg}, we formulate the problem of Imbalanced Regression, introducing $SERA$, the loss function that we will use to optimize our models. In addition, we also provide the details needed to embed this loss function in gradient boosting models, which will be our baselines. In Section~\ref{sec:gb}, we demonstrate how $SERA$ can be integrated as a custom optimization metric. In Section \ref{sec:exprstud}, we provide the experimental study and discuss the obtained results.
Finally, in Section \ref{sec:conc} we conclude our work with further research directions.

\section{Related Work} \label{sec:recentwork}

The study of imbalanced learning has been advocated over the years, as it poses well-known challenges to standard predictive learning tasks~\cite{imbsurvey}. There are three main strategies to cope with imbalanced domain learning problems: data-level, algorithm-level and hybrid. 

Data-level approaches are the most common ones. They allow for any standard machine learning to be used, as they act in a pre-processing stage by changing the data distribution to reduce the imbalance. Generally speaking, we can group them into under-sampling, over-sampling, generation of synthetic examples, or their combination.
Even though far more data-level methods have been proposed for classification, few exist for regression (e.g.~\cite{BRANCO201976}). An adaptation of the Synthetic Minority Over-sampling Technique (SMOTE)~\cite{smote}, initially proposed for classification, has been made for regression and named SMOTEr~\cite{smoter}. More recently, in the context of Deep Learning, a method to deal with missing data in an imbalanced regression domain was proposed in \cite{ddl}: Deep Imbalanced Regression (DIR). The specificity of this method lies in the fact that there may be missing values close to a high (low)-representative neighborhood in the target variable distribution. The distribution of the target variable is smoothed across the entire domain, considering a similarity kernel based on statistical properties of the data to estimate missing values. 
However, there is a caveat from these approaches -- they add artificial instances that may not represent the reality with which we are faced or remove common cases that can represent a crucial discriminating aspect for the predictive task.
Moreover, the models are not specifically optimized toward predicting those rare cases. Thus, it is not easy to assess whether the change made to the data distribution would effectively map to the intended predictive focus~\cite{nofreelunch}.

Regarding algorithm-level approaches, one of the most popular methods is cost-sensitive learning \cite{costsensboost,adacost,costsens}. These methods use costs to emphasize/relax errors committed by predictions at specific target values. An error committed at a rare or extreme value in imbalanced domains should have a higher cost than a common value. A problem linked to these methods is that assessing the exact cost of a given error is  highly domain-dependent and not straightforward~\cite{costsens}. 
Other methods include the optimization of an asymmetric loss function in standard learning algorithms \cite{asymNN,ehrig}. In the context of regression, few contributions have been made regarding optimization techniques. 
In~\cite{ehrig}, the authors focused on the prediction of extreme values by defining a branched asymmetric loss function in the residual space, using the Gradient Boosting algorithm as a training model. Here, the loss function branches into a quadratic function - for values around the mean, so-called normal; and exponential function - for extreme values. However, this loss function has the caveat of a precise definition of normality and thus the requirement of a  pre-defined threshold defined in the residual space. 

Recently, an ensemble model that consists of embedding SMOTE in several variations of the AdaBoost algorithm was proposed, both in classification \cite{smoteboost}, and regression tasks \cite{smoteboostr}. There is, however, a caveat to these models, as mentioned before - they add artificial instances that may not represent the reality with which we are faced.
This is especially critical when we are tackling ecological or health domains, where there is no guarantee that the generated instances may be valid observations. 

In standard regression tasks, a given model's quality or predictive power is typically assessed by metrics such as the Mean Squared Error ($MSE$) or the Mean Absolute Error ($MAE$). These metrics have one property in common: the importance attributed to each observation is uniform, which is not adequate if we are facing a problem of imbalanced regression. Several metrics were proposed for regression with the same goal of assigning uneven importance to instances. The Linear Exponential (LINEX)~\cite{linex} loss function, which, controlled by a parameter, differentiates over and underestimations. The Relevance-Weighted Root Mean Squared Error ($RW\text{-}RMSE$) \cite{nunophd}, a modified version of $RMSE$ which takes into the account the relevance of a given observation. This metric has the caveat of neglecting values which have a low relevance. The utility-based F-measure $F_\beta^u$ \cite{ribeiro}, is a function that depends on variations of both the well-known precision and recall, implemented in the context of regression. It relies on the values of relevance and utility assigned to a prediction for a given true value. Nevertheless, both values depend on a given threshold defined in the relevance and utility space.
All the metrics mentioned above share the same limitation: they are threshold dependent.

In this work, we follow the same principle as in~\cite{ehrig}, but with another loss function - Squared Error Relevance Area ($SERA$), a metric recently presented in the context of imbalanced regression~\cite{imbregex}.
Using the definition of relevance associated with the target variable domain, this metric explicitly gives the notion of asymmetry regarding the loss in different ranges of the target variable. This metric is not dependent on any threshold and thus does not face the problems referred to in the above metrics. Due to such characteristics, it presents the best option for exploring the possibility of model optimization in imbalanced regression tasks.

\section{Imbalanced Regression} \label{sec:imbreg}


Consider $\md$ a training set defined as $\md = \{\langle\boldsymbol{x}_i, y_i\rangle\}_{i=1}^N$, where $\boldsymbol{x}_i$ is a feature vector of the feature space $\mathcal{X}$ composed by $m$ independent variables and $y_i$ an instance of the feature space $\mathcal{Y}$ that depends on the feature space $\mathcal{X}$. In a supervised learning setting, our aim is to find the function $f$ that maps the feature space $\mathcal{X}$ onto $\mathcal{Y}$, $f: \mathcal{X} \to \mathcal{Y}$. Depending on the nature of $\mathcal{Y}$, we can face a classification (if $\mathcal{Y}$ is discrete) or a regression problem (if $\mathcal{Y}$ is continuous). To obtain the best approximation function of $f$, $h$, the standard approach in supervised learning is to consider a loss function $\ml$, responsible for the optimization of a set of parameters $\boldsymbol{\Theta}$ which tune a model to extract predictions that better describes new instances from the feature space $\mathcal{Y}$. 

Here, we will focus on the problem of imbalanced regression, i.e., when
the target variable $\mathcal{Y} \in \mathbb{R}$ presents a skewed distribution and the most important values for the prediction task are extreme (rare) values.
The most commonly used loss function in regression is the Mean Squared Error ($MSE$). However, this metric is not adequate for our prediction task.
The constant which minimizes $MSE$ is 
the mean of the target variable, $\bar{y}$,
which is counter intuitive for our predictive focus: the extreme values. 
In an imbalanced regression scenario, an appropriate loss function should search the parameter space $\boldsymbol{\Theta}$ such that it encompasses a good predictive power for both common (around the mean) and uncommon (extremes) instances of our target domain $\mathcal{Y}$. However, this is not a trivial task to accomplish.

\subsection{Relevance Function}\label{subsec:relev}

\begin{figure}[!hbt]
    \centering
    \includegraphics[width=\textwidth]{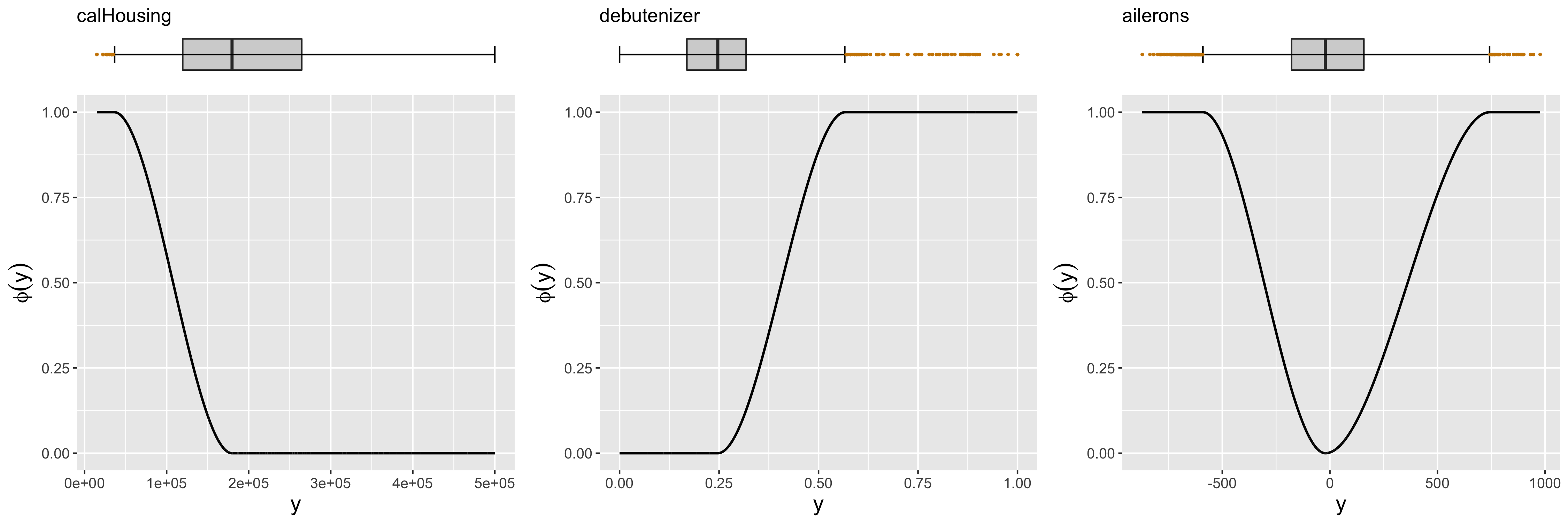}
    \caption{The adjusted boxplot (top) of a target variable $y$ and its automatically inferred relevance function $\phi(y)$ (bottom) for three data sets with different type of extreme values: low (left), high (middle) and both (right). 
    }
    \label{fig:adj_phi}
\end{figure}

In this study, we define an extreme value based on the notion of relevance introduced by~\cite{ribeiro}.
The authors define a relevance function $\phi: \mathcal{Y} \to [0,1]$ as a continuous function that expresses the application-specific bias concerning the target variable domain $\mathcal{Y}$ by mapping it into a $[0,1]$ scale of relevance, where 0 and 1 represent the minimum and maximum relevance, respectively. 
With the assumption that extreme values are the values of interest, authors have also proposed a method that automatically constructs the $\phi(.)$ function. 
It achieves that by interpolating a set of control points provided by the adjusted boxplot, a non-parametric modification to Tukey's boxplot, proposed by \cite{boxbox}.
In particular, this method uses the median and the whiskers as the set of key points to interpolate. 
In Figure \ref{fig:adj_phi} we depict the result of the automatic mapping of the 
adjusted boxplot to the relevance space, as introduced by~\cite{ribeiro}, for three different scenarios based on the type of extremes indicated to be of interest: low, high or both (default).

\subsection{Squared Error Relevance Area (SERA)}\label{subsec:sera}

Once we have the domain of the target variable mapped into a relevance space, we now present the asymmetric loss function $SERA$. We will use $SERA$ to improve the predictive power of extreme values in Gradient Boosting algorithms, emphasizing that any model which relies on optimization in the parameter space $\boldsymbol{\Theta}$ could be used to perform this improvement.

Let $\mathcal{D} = \{\langle\boldsymbol{x}_i, y_i\rangle\}_{i=1}^N$ be a data set and $\phi: \mathcal{Y} \to \{0,1\}$ a relevance function defined for the target variable $Y$. Considering the subset $\md^t \subseteq \mathcal{D}$ of instances such that $\md^t = \{\langle \boldsymbol{x}_i, y_i\rangle \in \mathcal{D}\, |\, \phi(y_i) \geq t\}$, we can define a Squared Error-Relevance ($SER_t$) to estimate the error of a given model with respect to a given cutoff $t$ as

\beq SER_t = \sum_{y_i \in \md^t} (\yh_i - y_i)^2~ \eeq

This is the sum of squared errors for all the instances such that the relevance of the target value is bounded by a given threshold $t$.
Since this metric only depends on instances such that $\phi(y_i) \geq t$, we will have that, for any given $\delta \in \mathbb{R}^+$, s.t. $t + \delta \leq 1$: 
$SER_{t+\delta} \leq SER_{t}$. 
Finally, its maximum and minimum value are ascertained when $t = 0$ and $t = 1$, respectively.

In the same work, the authors took a step further and integrated this estimate w.r.t. all possible cutoff values (i.e., between 0 and 1). There, they defined this area as the Squared Error Relevance Area ($SERA$), and it is given by

\beq \label{eq:sera} SERA = \int_{0}^{1} SER_t~dt = \int_{0}^{1} \sum_{y_i \in \md^t}(\hat{y}_{i} - y_i)^2 ~ dt \eeq

This area has some important properties that can help us better understand the performance of models in an imbalanced regression setting. 
First, it encompasses all the possible relevance thresholds constrained in the definition of $SER_t$, removing the need to explicitly define a threshold. Secondly, it is a decreasing and monotonic function. Note that we are integrating along with all possible relevance values. Since we know by definition that $\md^{t + \delta} \subseteq \md^t$, the higher the relevance threshold is, the lower will be the number of instances considered, but also more relevant. Thus, on the one hand, values of $SER_t$ which have a high relevance will have a greater contribution to this area when compared with instances where the relevance is small. The squared errors of these latter instances are accounted less times for $SERA$, when compared to the high relevance instances. 
On the other hand, the area will be smaller at points where the relevance is high (since we are only considering the observations that have a high relevance). 
In this sense, we are explicitly penalizing high relevance errors, which are usually harder to optimize while keeping the entire data domain. Finally, since this metric is built by integrating $SER_t$ over all the relevance domain, which is convex, convexity is also preserved. Also, this metric must be differentiable. By the same token, given that $SER_t$ is differentiable, so is $SERA$.

\section{Optimization Loss Function for Imbalanced Regression} \label{sec:gb}

We aim to study the possibility and impact of embedding $SERA$ as an optimization loss function in supervised learning algorithms. Its use in already implemented learning algorithms only requires the proposal of a custom loss function, where the only thing we need to provide is the first and second-order derivatives. 

The first-order derivative of $SERA$ is obtained by evaluating the first derivative w.r.t. a given prediction $\hat{y}_j$, as follows

\beq \begin{aligned}
    \frac{\partial SERA}{\partial \yh_j} &= \frac{\partial}{\partial \yh_j}\int_0^1 \sum_{y_i \in \md^t}(\hat{y}_{i} - y_i)^2~dt \\
    &= 2 \int_0^1 \sum_{y_i \in \md^t}(\hat{y}_{i} - y_i)~\delta_{ij}~dt
\end{aligned} \eeq 

\noindent where $\delta_{ij}$ is the Kronecker's delta, which takes value of 1 if $i = j$ and 0 otherwise. Since we need to take into the account all the possible relevance values a given observation is encompassed in, we can write the expression above as

\beq \label{eq:sera_deriv_trap} \frac{\partial SERA}{\partial \yh_j} = 2 \int_0^1 (\hat{y}_j - y_j)\bigg|_{y_j \in \md^t} ~ dt\eeq

The second derivative w.r.t. a given prediction $\hat{y}_j$ is obtained by

\beq \label{eq:sera_dderiv_trap} \frac{\partial^2 SERA}{\partial \yh_j^2} = 2 \int_0^1 \boldsymbol{1}(y_j \in \md^t)~dt\eeq

\noindent where $\boldsymbol{1}(.)$ is an indicator which takes the value of $1$ if the argument holds, and $0$ otherwise. 

In this paper, we will drive our efforts using Gradient Boosting algorithms, namely two well-known variants of Gradient Boosting Regression Trees (GBRT) \cite{gradboost}, XGBoost \cite{xgboost} and LGBM \cite{lightgbm}.
For that, we resort to the implementations found in \textbf{R}~\cite{rsoft} from the packages \texttt{xgboost}~\cite{xgboostR} and \texttt{lightgbm}~\cite{lgbmR}, respectively.

To approximate the values of the two derivatives, 
we resort to the trapezoidal rule with a uniform grid of $T$ equally spaced intervals between $[0, 1]$. In this approximation, we can expand the summations and deduce the following expressions (see Appendix \ref{appendix:sera_app} for all derivation steps).

\begin{align} \label{derivs_cmp} 
\frac{\partial SERA}{\partial \yh_j} &\approx \frac{1}{T} \left(1 + 2 n_j +  \boldsymbol{1}\left(y_j \in \md^{t_T} \right) \right) (\yh_j - y_j)~~, \nonumber \\ \frac{\partial^2 SERA}{\partial \yh_j^2} &\approx \frac{1}{T} \left(1 + 2 n_j +  \boldsymbol{1}\left(y_j \in \md^{t_T} \right) \right) 
\end{align}

\noindent where $n_j \in [1, T-1]$ is the number of times the instance $y_j$ contributes to $SERA$ derivative. From this, we can infer that, for a given prediction, the first and second-order derivatives will be greater (assuming a greater error for extreme values) as the relevance increases, as there will be a higher contribution from $n_j$. 
These derivatives, in addition with $SERA$, were implemented in \textbf{R}~\cite{rsoft}.

From now on, we will designate XGBoost and LGBM models optimized with $SERA$ as \xgbs and \lgbms, while models optimized with $MSE$ \xgbm and \lgbmm, respectively.

\subsubsection{Computational Complexity}

Another important aspect is the computational complexity introduced by $SERA$.
The trapezoidal rule has a computational complexity of $O(T)$, where $T$ is the number of steps taken to discretize an integral. $SER_t$ has a computational complexity of $O(|\mathcal{D}^t|)$, where $|\mathcal{D}^t|$ is the number of instances with relevance higher or equal to a given threshold $t$. 
$SERA$ will consider $|\mathcal{D}^{t_0}| + |\mathcal{D}^{t_1}| + ... + |\mathcal{D}^{t_T}|$ instances for all the $T$ steps of the Riemann's sum. 
In the worst-case scenario, all the target values have a constant and maximum relevance equal to 1. In that case, $|\mathcal{D}|$ is the number of instances accounted for all steps.
Thus, $SERA$ will have a computational complexity of $O(T \times |\mathcal{D}|)$. 
Regarding the computational complexity introduced in XGBoost and LGBM, we only need to take into consideration the additional complexity of the first and second order derivatives. 
Thus, using the approximation found in Appendix \ref{appendix:sera_app}, 
the computational complexity 
will be again $O(T \times |\mathcal{D}|)$.

\section{Experimental Study} \label{sec:exprstud}

Our goal is to answer the research question (\textbf{Q}) that motivated this work: can $SERA$ be used as an optimization loss function to reduce errors for both extreme and common values?

In this section, we take into consideration a group of 36 regression data sets from several domains  
with an imbalanced distribution on the target variable. Given these data sets and the models described in Section \ref{sec:gb}, we will start by describing our experimental setup in Section \ref{subsec:expsetup}. Namely, the considered data sets with an imbalanced domain, and the grid-search procedure for parameter tuning of models. In Section \ref{sub:modeloptim} we refer to the Bayes Sign Test used to assess the statistical significance of the results. Given the best parameters for each model, we present and discuss the obtained results for all data sets in Section \ref{sub:res_oos}.

\subsection{Experimental Setup} \label{subsec:expsetup}

To study the effects of using $SERA$ as an optimization loss function, a wide range of data sets from several domains in the context of imbalanced regression is used. These data sets, with their respective main properties, are presented in Table~\ref{tab:data sets}. From them, we extracted the number of instances $|\md|$, the number of nominal (Nom) and numerical (Num) variables. In addition, and to give a notion of the imbalance present in the target variable, we resort to the automatic method proposed in ~\cite{imbregex} that is based on the adjusted boxplot.
We calculated the number of instances such that $\phi(y) = 1$, as representative of the number of extreme (rare) target value instances $(|\md_R|)$ and the Imbalance Ratio ($IR$), calculated as $|\md_R|/|\md| \times 100\%$. Finally, we also include the type of imbalance for each target variable as follows: if the adjusted boxplot only presents outliers below or above the respective fence, the type of extremes is low (L) or high (H), respectively, while if it presents outliers below and above the fences, the type is both (B).

\begin{table}
\caption{Data sets description: $|\md|$ - nr of instances, \textbf{Nom} - nr. of nominal attributes, \textbf{Num} - nr. of numeric attributes,  $|\md_R|$ - nr. of extreme (rare) instances, i.e. $\phi(y) = 1$ $IR$ - imbalance ratio and Type - type of extremes.}
\label{tab:data sets}
\scriptsize
\begin{minipage}{0.49\linewidth}
\resizebox{\textwidth}{!}{
\begin{tabular}{cccccccc}
\toprule
\textbf{id} & \textbf{dataset} & $|\md|$ & \textbf{Nom} & \textbf{Num} & $|\md_R|$ & $IR$ & Type \\
\midrule
1 & diabetes & 35 & 0 & 3 & 4 & 12.90 & H \\ 
2 & triazines & 151 & 0 & 61 & 4 & 2.72 & B \\ 
3 & a7 & 160 & 3 & 9 & 7 & 4.58 & H \\ 
4 & autoPrice & 165 & 10 & 16 & 3 & 1.85 & L \\ 
5 & elecLen1 & 399 & 0 & 3 & 4 & 1.01 & H \\ 
6 & housingBoston & 407 & 0 & 14 & 40 & 10.90 & B \\ 
7 & forestFires & 416 & 0 & 13 & 7 & 1.71 & H \\ 
8 & wages & 429 & 7 & 4 & 1 & 0.23 & B \\ 
9 & strikes & 501 & 0 & 7 & 1 & 0.20 & H \\ 
10 & mortgage & 841 & 0 & 16 & 60 & 7.68 & L \\ 
11 & treasury & 841 & 0 & 16 & 79 & 10.37 & L \\ 
12 & musicorigin & 848 & 0 & 118 & 15 & 1.80 & B \\ 
13 & airfoild & 1203 & 0 & 6 & 11 & 0.92 & H \\ 
14 & acceleration & 1387 & 3 & 12 & 30 & 2.21 & B \\ 
15 & fuelConsumption & 1413 & 12 & 26 & 27 & 1.95 & B \\ 
16 & availablePower & 1443 & 7 & 9 & 75 & 5.48 & B \\ 
17 & maxTorque & 1442 & 13 & 20 & 43 & 3.07 & B \\ 
18 & debutenizer & 1918 & 0 & 8 & 90 & 4.92 & H \\
\bottomrule
\end{tabular}
}
\end{minipage} 
\begin{minipage}{0.52\linewidth}
\resizebox{\textwidth}{!}{
\begin{tabular}{cccccccc}
\toprule
\textbf{id} & \textbf{data set} & $|\md|$ & \textbf{Nom} & \textbf{Num} & $|\md^R|$ & $IR$ & Type \\
\midrule
  19 & space\_ga & 2487 & 0 & 7 & 21 & 0.85 & B \\ 
  20 & pollen & 3080 & 0 & 5 & 32 & 1.05 & B \\ 
  21 & abalone & 3343 & 1 & 8 & 374 & 12.60 & B \\ 
  22 & wine & 5199 & 0 & 12 & 1022 & 24.47 & H \\ 
  23 & deltaAilerons & 5705 & 0 & 6 & 528 & 10.20 & B \\ 
  24 & heat & 5922 & 3 & 9 & 39 & 0.66 & B \\ 
  25 & cpuAct & 6555 & 0 & 22 & 227 & 3.59 & L \\ 
  26 & kinematics8fh & 6556 & 0 & 9 & 50 & 0.77 & B \\ 
  27 & kinematics32fh & 6556 & 0 & 33 & 53 & 0.82 & B \\ 
  28 & pumaRobot & 6556 & 0 & 33 & 91 & 1.41 & B \\ 
  29 & deltaElevation & 7615 & 0 & 7 & 1802 & 31 & H \\ 
  30 & sulfur & 8065 & 0 & 6 & 606 & 8.12 & B \\ 
  31 & ailerons & 11003 & 0 & 41 & 186 & 1.72 & B \\ 
  32 & elevators & 13280 & 0 & 18 & 1598 & 13.68 & B \\ 
  33 & calHousing & 16513 & 0 & 9 & 23 & 0.14 & L \\ 
  34 & house8H & 18229 & 0 & 9 & 305 & 1.70 & B \\ 
  35 & house16H & 18229 & 0 & 17 & 303 & 1.69 & B \\ 
  36 & onlineNewsPopRegr & 31716 & 0 & 60 & 2879 & 9.98 & B \\ 
\bottomrule
\end{tabular}
}
\end{minipage}
\end{table}

To assess the effectiveness of each model, we 
performed a random partition for each data set, where 80\% will be used to tune the models while the remaining 20\% to make predictions under the best model configuration found in a given data set. To tune the parameters for each model, we will use a grid-search approach with a $10$-fold stratified cross-validation. We define a workflow of a given algorithm $j$ as the tuple $\boldsymbol{W}^{(j)} = (\boldsymbol{M}_j, \boldsymbol{\Theta}^{(j)}) = \{\mw_q^{(j)}\}_{q=1}^e$, where $e$ is the number of different workflows considered for a given tuple, $\boldsymbol{M}_j$ denote the algorithm used, $\boldsymbol{\Theta}^{(j)}$ the respective set of parameters, which are described in Table \ref{tab:parameters_grid}.
\begin{table}[!hbt]
    \centering
    \scriptsize
    \caption{Models parameters considered for grid search.}
    \begin{tabular}{lll}
    \toprule
    \textbf{Model} & \textbf{R Package} & \textbf{Parameters} \\
    \midrule
    \multirow{3}{5em}{XGBoost LGBM} & \multirow{3}{8em}{xgboost~\cite{xgboostR} lightgbm~\cite{lightgbm}} & 
    \multirow{3}{11em}{nrounds = $\{250, 500\}$
                       max\_depth = $\{3, 5, 7\}$ 
                       $\eta = \{10^{-3}, 10^{-2}, 10^{-1}\}$} \\
    & & \\
    & & \\
    \bottomrule
    \end{tabular}
    \label{tab:parameters_grid}
\end{table}

Given the workflows obtained from the grid-search, we start by providing a methodology to answer the question that motivated this work, \textbf{Q}. It consists on the following tasks.

\begin{enumerate}
    \item[\textbf{T1}:] For each data set, and for each model in $\boldsymbol{M}$, we select the workflow that had the lowest score according to $SERA$. This score is calculated by averaging the results obtained by cross-validation on the 80\% partition.
    \item[\textbf{T2}:]  Given the best workflows, we compare them using the Bayes Sign Test \cite{bayes} (Section \ref{sub:modeloptim}). 
    The designation we give to Gradient Boosting models is $\mw^M$, in case they are optimized using a standard loss function ($MSE$) and $\mw^S$ in case they are optimized using $SERA$.
    \item[\textbf{T3}:]  Next, we train our best workflows for each data set with the partitioned $80\%$ and, with the remaining $20\%$, we assess the quality of their predictions (Section \ref{sub:res_oos}). This quality will also be evaluated by plotting $SERA$ curves.
\end{enumerate}

\subsection{Results on Model Optimization}
\label{sub:modeloptim}

With the top workflows from each model obtained by \textbf{T1}, we can assess the performance of our models in task \textbf{T2}. For that, we resort to the Bayes Sign Test. Briefly, this test compares two models on a multi data set scenario by measuring their score difference (a prior probability) for all data sets, returning a probability measure (the posterior) hinting if a model is practically better than another, or if they are equivalent. This equivalence is measured in a given interval and is defined as the Region Of Practical Equivalence (ROPE) \cite{rope}. 
The prior $z_i$, where $i$ indicates a given data set, is determined by averaging the normalized difference below for all $k$-folds

\beq \label{eq:priors} z_i = \frac{1}{10} \sum_{k=1}^{10} \frac{\ml_k(\mw^S) - \ml_k(\mw^M)}{\ml_k(\mw^M)}~~, \eeq

and taking $\ml$ as $SERA$ or $MSE$. After determining this mean difference for all data sets, we feed into the Bayes Sign Test the vector $\boldsymbol{z}$ and a ROPE between $[-1\%, 1\%]$, returning the posterior probability $p(z)$ that a given model is practically better or equivalent than the other. 

\begin{figure}[!hbt]
    \centering
    \includegraphics[width=.6\textwidth]{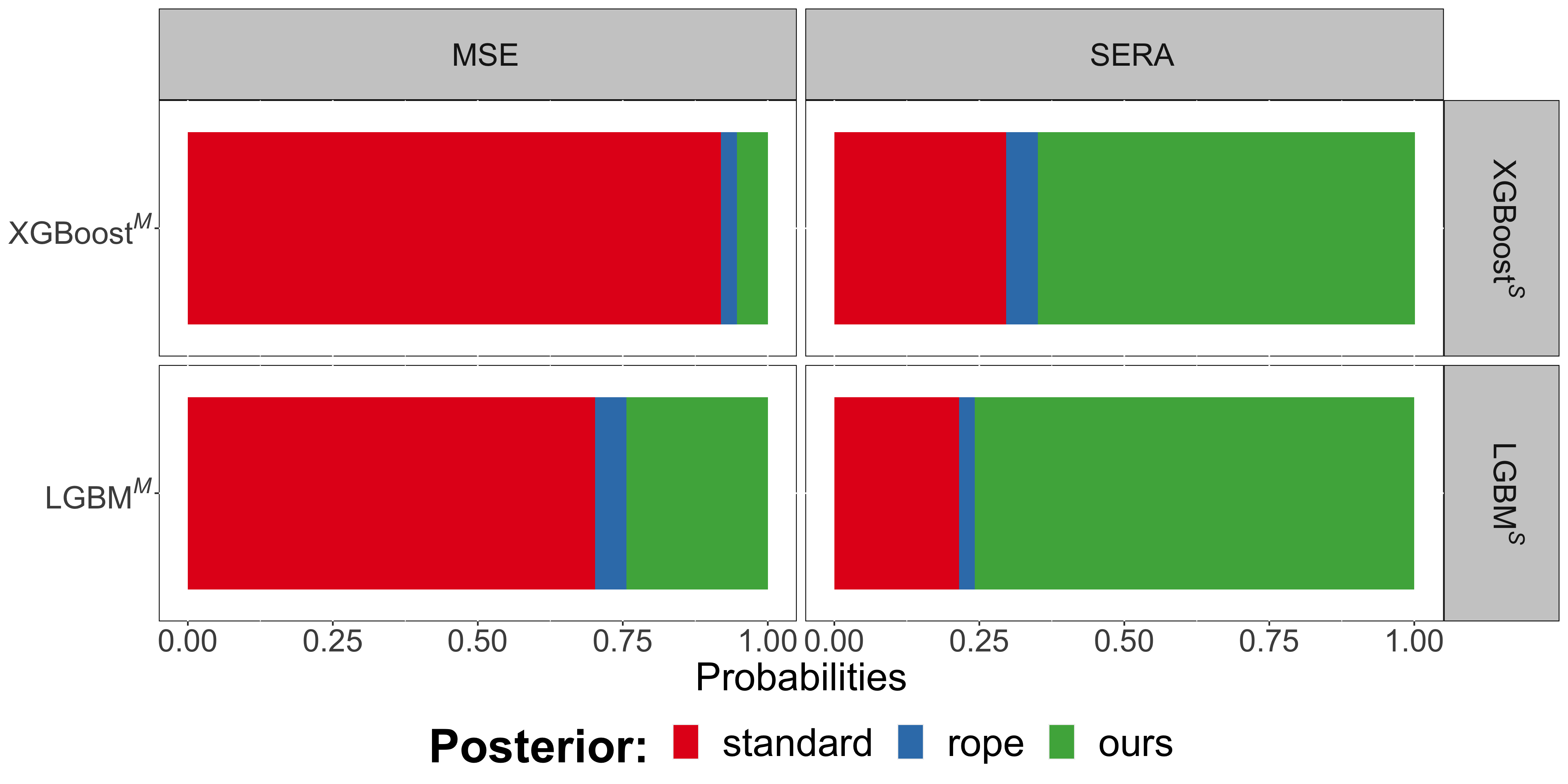}
    \caption{Comparison between our models optimized with $SERA$, \lgbms and \xgbs, against the standard models \lgbmm and \xgbm. Each color denotes the probability of our implementation (green) or standard (red) being practically better or equivalent (blue) to one another according with the Bayes Sign Test with the ROPE interval $[-1\%,1\%]$. The left and right column denote the results 
    of the Bayes Sign Test with $MSE$ and $SERA$, respectively.}
    \label{fig:bayes_results}
\end{figure}

The results from this evaluation are depicted in Figure \ref{fig:bayes_results} and provide us with two perspectives according to the considered error metrics.
Regarding $MSE$ (left column), the standard models are practically better (with a $p(z)$ of 0.92 for \xgbm and a $p(z)$ of 0.7 for \lgbmm). 
Concerning $SERA$ (right column), results tell us that both algorithms with our optimization are practically better (with a $p(z)$ of 0.65 for \xgbs and a $p(z)$ of 0.76 for \lgbms).


Thus, from a statistical points of view and in a model optimization scenario, there is a clear trade-off between the standard and our models when assessing their scores with different metrics. This was somewhat expected as these metrics have a different predictive focus as it was already mentioned above. Nevertheless, from this test we are able to infer the ability of $SERA$ as a loss function to lower the errors obtained in a problem of Imbalance Regression. With this, we finish our second task \textbf{T2} and partially answered our main question \textbf{Q}. Next, we aim to show that $SERA$ do in fact improve the predictive power for both common and extreme values in an out-of-sample scenario (i.e., using our test data).

\subsection{Results in Out-of-Sample}\label{sub:res_oos}

Using the parameters found in the best workflows obtained for each model and for each data set in task \textbf{T1}, we train our models and assess their predictions with the ($20\%$) out-of-sample data. Given these predictions, we calculated $SERA$ and $MSE$ for all the models in the considered data sets (cf. Tables~\ref{tab:results2} and \ref{tab:results1} in Appendix~\ref{appendix:results}). 

\begin{figure}[!hbt]
    \centering
    \includegraphics[width=.6\textwidth]{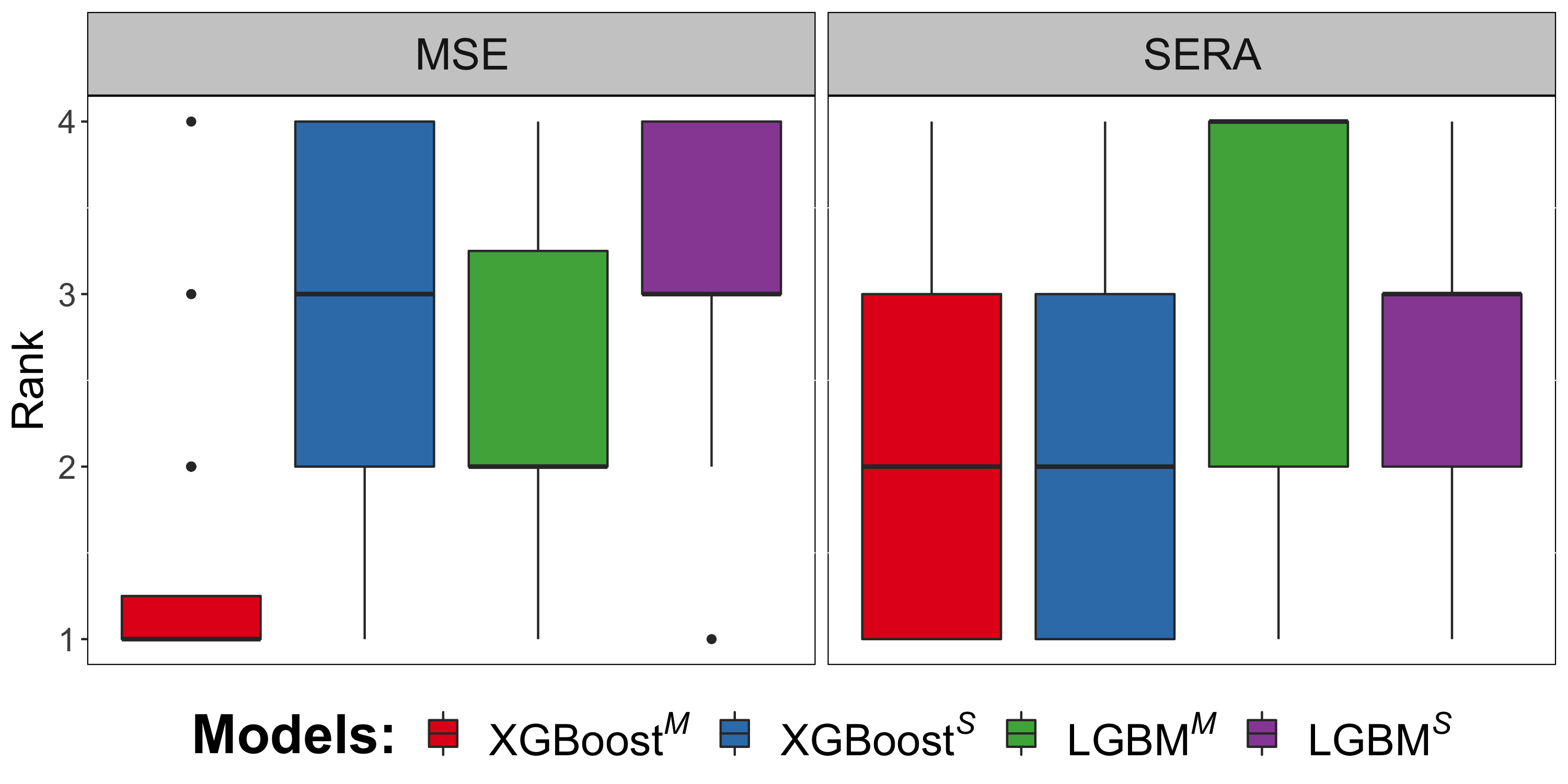}
    \caption{Rank distribution of models by $MSE$ and $SERA$ results in out-of-sample.}
    \label{fig:rank}
\end{figure}

From the obtained results, we initially take into consideration a rank evaluation of our models. For that purpose, and for a given data set, the rank of 1 is assigned to the model which provided the lowest score. Figure \ref{fig:rank} depicts the rank distribution for each model over all the considered data sets. 
In $MSE$-based ranking (left column), \xgbm~ had a median rank of 1, followed by \lgbmm~ with a rank of 2, and finally \xgbs~ and \lgbms, both with a median rank of 3. In $SERA$-based ranking (right column), both \xgbm and \xgbs~ had a median rank of two, followed by \lgbms~ with a median rank of 3, and finally \lgbmm~ with a rank of 4.
From these ranking distributions, we can take a somewhat expected conclusion: standard models were the top performers when evaluated under $MSE$, and an unexpected one: under $SERA$, \xgbm~ and \xgbs~ compete with each other.

Next, we focus on each algorithm independently. 
While for XGBoost, \xgbs\ ~~had the lowest score in 3 and 18 of the data sets when evaluated by $MSE$ and $SERA$, respectively, 
for LGBM, \lgbms ~~had a better score in 11 and 26 of the data sets when evaluated by $MSE$ and $SERA$.

\begin{figure}[!hbt]
    \centering
    \includegraphics[width=\textwidth]{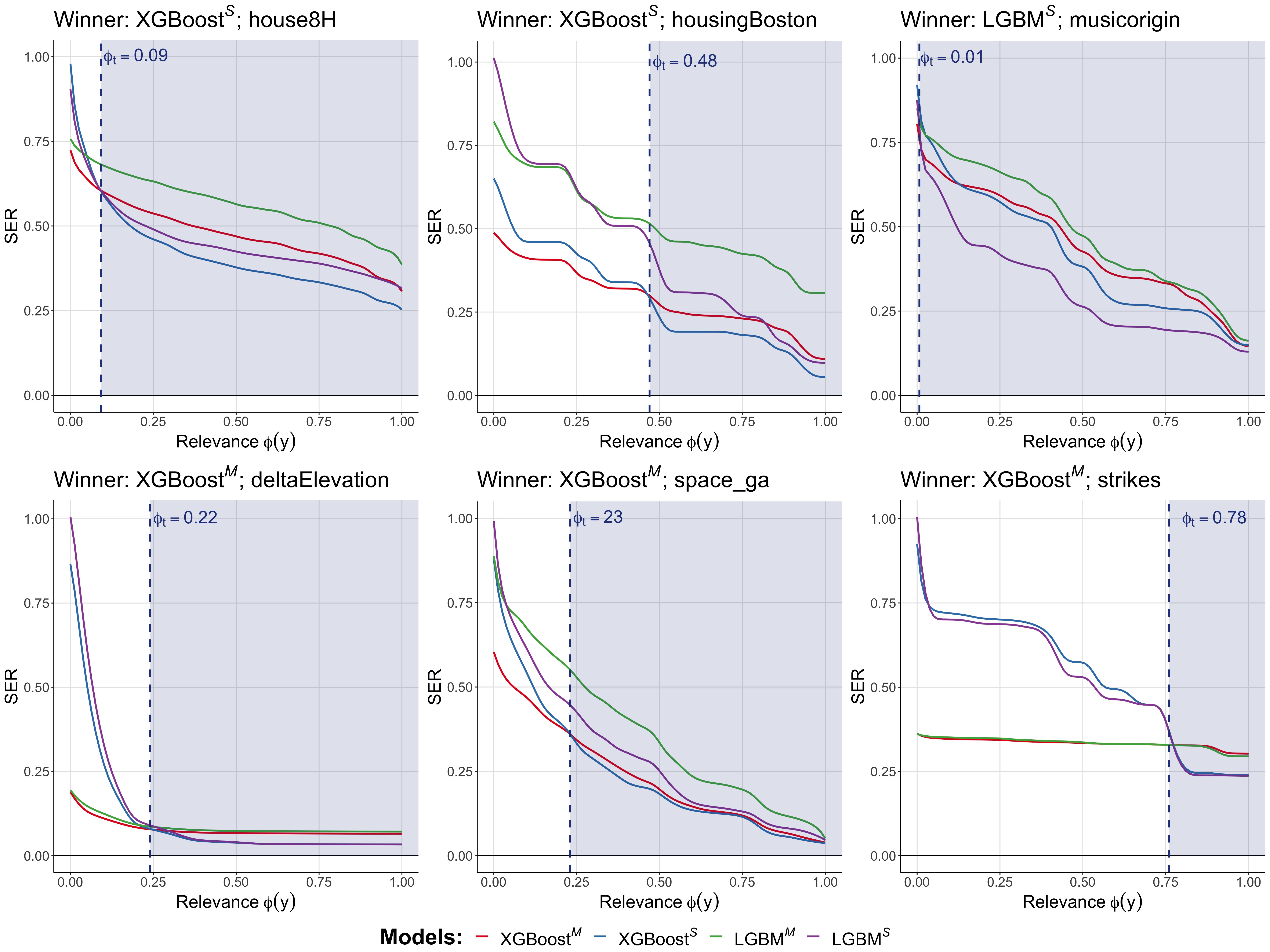}
    \caption{$SERA$ curves for six selected data sets. The first row provides data sets where our models had the lowest error, while the second row provides data sets where \xgbm~ had the lowest error. The highlighted region in each graph depicts the turning point where models optimized with $SERA$ started to have a lower error w.r.t. standard models.}  
    \label{fig:sera_plots}
\end{figure}

Although the standard models had the lowest score in more data sets than ours, it does not mean that, for a given data set, those errors were the lowest across all the relevance domain. To study this statement, we follow \cite{imbregex} and show $SERA$ curves for six selected data sets. These curves are built by calculating the error $SER_t$ as the relevance threshold $t$ for $\phi(y)$ increases and are shown in Figure \ref{fig:sera_plots}. 

In addition to $SERA$ curves, we also define a turning point for each data set. That point is the minimum relevance value $\varphi$ for which a model optimized with $SERA$ 
has a $SERA$ estimate for all the values with a relevance greater or equal to $\varphi$, i.e. $SERA_{\phi(.) \geq \varphi}(\mm^S)$,
lower than the $SERA$ estimate obtained by the standard model 
in the same conditions, i.e. for all the values with a relevance greater or equal to $\varphi$, i.e. $SERA_{\phi(.) \geq \varphi}(\mm^M)$.
More formally, and for a specific data set, the turning point is then a threshold $\phi_t$ obtained by \begin{equation}
\label{eq:turningpoint}
\phi_t = \text{min}\{\varphi \in [0, 1] ~ | ~ SERA_{\phi(.) \geq \varphi}(\mm^S) < SERA_{\phi(.) \geq \varphi}(\mm^M)\}.
\end{equation}
In the curves of Figure \ref{fig:sera_plots}, the turning points are represented by dashed lines, and the shadowed regions represent the relevance domain for which the condition above holds.


The plots from the first row of Figure \ref{fig:sera_plots} show us $SERA$ curves where models that were optimized with $SERA$ had the lowest score. As depicted, for house8H data set, \xgbs~ had the lowest score in most of the relevance domain ($\phi_t = 0.09$). For housingBoston data set, although \xgbs~ has the lowest $SERA$, its turning point occurs at a higher threshold ($\phi_t = 0.48)$. 
Finally, we show the curves for 
musicorigin data set for which \lgbms~ had the lowest score. Here, the turning point also occurs at a very low relevance ($\phi_t = 0.01$). 

Regarding the second row of plots of Figure \ref{fig:sera_plots}, we show three data sets where \xgbm~ had the lowest $SERA$ score. From these examples, we see that although \xgbm~ was the top performer, there are several turning points in the relevance domain for which our models surpassed \xgbm~($\phi_t=0.22$, $\phi_t = 0.23$, $\phi_t = 0.78$ for deltaElevation, space\_ga and strikes data sets, respectively). 

From this analysis we can conclude that $SERA$ can be used as a loss function to reduce errors in both extreme and common values (first row of Figure \ref{fig:sera_plots}). Even when models optimized with $SERA$ do not provide the best score across the whole relevance domain, they can still perform better for different relevance domains (second row from Figure \ref{fig:sera_plots}). With this, we conclude our task \textbf{T3} and answered the question that motivated this work \textbf{Q}.

\section{Conclusions} \label{sec:conc}

In this work, we addressed the problem of embedding $SERA$ as an optimization loss function to improve the predictive power for both extreme and normal values. We used as our baselines two well-known variations of Gradient Boosting models, XGBoost and LGBM.
Results showed that the embedment of $SERA$ as an optimization loss function in our algorithms provided the ability of reducing errors for both extreme and normal values. In addition, when underperformed by standard models, our algorithms were able to provide a more accurate prediction in a high relevance domain. For the sake of reproducibility, all the experiments are available in github\footnote{\href{https://github.com/anibalsilva1/IRModelOptimization}{https://github.com/anibalsilva1/IRModelOptimization}}.

Finally, we highlight that hyper-parameter optimization, due to the construction of $SERA$, is not restricted to the considered algorithms, opening the use of this loss function to several domains of Machine Learning (e.g. Deep Learning) in the realm of Imbalanced Regression.

\section*{Acknowledgements}
This work was supported by the CHIST-ERA grant CHIST-ERA-19-XAI-012, and project CHIST-ERA/0004/2019 funded by FCT.

\bibliographystyle{splncs04}
\bibliography{bibdb}

\begin{thebibliography}{10}
\providecommand{\url}[1]{\texttt{#1}}
\providecommand{\urlprefix}{URL }
\providecommand{\doi}[1]{https://doi.org/#1}

\bibitem{bayes}
Benavoli, A., Corani, G., Dem{\v{s}}ar, J., Zaffalon, M.: Time for a change: a
  tutorial for comparing multiple classifiers through bayesian analysis.
  Journal of Machine Learning Research  \textbf{18}(77),  1--36 (2017)

\bibitem{imbsurvey}
Branco, P., Torgo, L., Ribeiro, R.P.: A survey of predictive modeling on
  imbalanced domains. ACM Comput. Surv.  \textbf{49}(2) (2016).
  \doi{10.1145/2907070}

\bibitem{BRANCO201976}
Branco, P., Torgo, L., Ribeiro, R.P.: Pre-processing approaches for imbalanced
  distributions in regression. Neurocomputing  \textbf{343},  76--99 (2019).
  \doi{10.1016/j.neucom.2018.11.100}

\bibitem{smote}
Chawla, N.V., Bowyer, K.W., Hall, L.O., Kegelmeyer, W.P.: Smote: Synthetic
  minority over-sampling technique. J. Artif. Int. Res.  \textbf{16}(1),
  321–357 (2002)

\bibitem{smoteboost}
Chawla, N.V., Lazarevic, A., Hall, L.O., Bowyer, K.W.: Smoteboost: Improving
  prediction of the minority class in boosting. In: Knowledge Discovery in
  Databases: PKDD 2003. pp. 107--119. Springer Berlin Heidelberg, Berlin,
  Heidelberg (2003)

\bibitem{xgboost}
Chen, T., Guestrin, C.: Xgboost: A scalable tree boosting system. In: 22nd ACM
  SIGKDD International Conference on Knowledge Discovery and Data Mining. p.
  785–794. KDD '16, ACM (2016). \doi{10.1145/2939672.2939785}

\bibitem{xgboostR}
Chen, T., He, T., Benesty, M., Khotilovich, V., Tang, Y., Cho, H., Chen, K.,
  Mitchell, R., Cano, I., Zhou, T., Li, M., Xie, J., Lin, M., Yifeng, G., Li,
  Y., Yuan, J.: xgboost: Extreme Gradient Boosting (2022),
  \url{https://CRAN.R-project.org/package=xgboost}

\bibitem{linlin}
Christoffersen, P.F., Diebold, F.X.: Further results on forecasting and model
  selection under asymmetric loss. Journal of Applied Econometrics
  \textbf{11}(5),  561--571 (1996)

\bibitem{ehrig}
Ehrig, L., Atzberger, D., Hagedorn, B., Klimke, J., Döllner, J.: Customizable
  asymmetric loss functions for machine learning-based predictive maintenance.
  In: 2020 8th International Conference on Condition Monitoring and Diagnosis
  (CMD). pp. 250--253 (2020). \doi{10.1109/CMD48350.2020.9287246}

\bibitem{costsens}
Elkan, C.: The foundations of cost-sensitive learning. 17th International
  Conference on Artificial Intelligence  \textbf{1},  973--978 (2001)

\bibitem{adacost}
Fan, W., Stolfo, S.J., Zhang, J., Chan, P.K.: Adacost: Misclassification
  cost-sensitive boosting. In: 16th International Conference on Machine
  Learning. p. 97–105. ICML '99, Morgan Kaufmann Publishers Inc. (1999)

\bibitem{gradboost}
Friedman, J.H.: Greedy function approximation: A gradient boosting machine. The
  Annals of Statistics  \textbf{29}(5),  1189--1232 (2001),
  \url{http://www.jstor.org/stable/2699986}

\bibitem{clive}
Granger, C.W.J.: Outline of forecast theory using generalized cost functions.
  Spanish Economic Review  \textbf{1}(2),  161--173 (1999).
  \doi{10.1007/s101080050007}

\bibitem{boxbox}
Hubert, M., Vandervieren, E.: An adjusted boxplot for skewed distributions.
  Computational Statistics \& Data Analysis  \textbf{52},  5186--5201 (2008).
  \doi{10.1016/j.csda.2007.11.008}

\bibitem{lightgbm}
Ke, G., Meng, Q., Finley, T., Wang, T., Chen, W., Ma, W., Ye, Q., Liu, T.Y.:
  Lightgbm: A highly efficient gradient boosting decision tree. In: 31st
  International Conference on Neural Information Processing Systems. p.
  3149–3157. NIPS'17, Curran Associates Inc. (2017)

\bibitem{rope}
Kruschke, J., Liddell, T.: The bayesian new statistics: Two historical trends
  converge. SSRN Electronic Journal  (2015). \doi{10.2139/ssrn.2606016}

\bibitem{nunophd}
Moniz, N.: Prediction and Ranking of Highly Popular Web Content. Ph.D. thesis,
  Faculty of Sciences - University of Porto (2017)

\bibitem{nofreelunch}
Moniz, N., Monteiro, H.: No free lunch in imbalanced learning. Knowledge-Based
  Systems  \textbf{227},  107222 (2021). \doi{10.1016/j.knosys.2021.107222}

\bibitem{smoteboostr}
Moniz, N., Ribeiro, R., Cerqueira, V., Chawla, N.: Smoteboost for regression:
  Improving the prediction of extreme values. In: IEEE 5th International
  Conference on Data Science and Advanced Analytics (DSAA). pp. 150--159
  (2018). \doi{10.1109/DSAA.2018.00025}

\bibitem{rsoft}
{R Core Team}: R: A Language and Environment for Statistical Computing. R
  Foundation for Statistical Computing, Vienna, Austria (2020),
  \url{https://www.R-project.org/}

\bibitem{asymNN}
Rengasamy, D., Rothwell, B., Figueredo, G.P.: Asymmetric loss functions for
  deep learning early predictions of remaining useful life in aerospace gas
  turbine engines. In: International Joint Conference on Neural Networks
  (IJCNN). pp.~1--7 (2020). \doi{10.1109/IJCNN48605.2020.9207051}

\bibitem{imbregex}
Ribeiro, R., Moniz, N.: Imbalanced regression and extreme value prediction.
  Machine Learning  \textbf{109},  1--33 (2020).
  \doi{10.1007/s10994-020-05900-9}

\bibitem{lgbmR}
Shi, Y., Ke, G., Soukhavong, D., Lamb, J., Meng, Q., Finley, T., Wang, T.,
  Chen, W., Ma, W., Ye, Q., Liu, T.Y., Titov, N.: lightgbm: Light Gradient
  Boosting Machine (2022), \url{https://CRAN.R-project.org/package=lightgbm}

\bibitem{costsensboost}
Sun, Y., Kamel, M.S., Wong, A.K., Wang, Y.: Cost-sensitive boosting for
  classification of imbalanced data. Pattern Recognition  \textbf{40}(12),
  3358--3378 (2007). \doi{10.1016/j.patcog.2007.04.009}

\bibitem{smoter}
Torgo, L., Ribeiro, R.P., Pfahringer, B., Branco, P.: Smote for regression. In:
  Progress in Artificial Intelligence. pp. 378--389. Springer (2013)

\bibitem{ribeiro}
Torgo, L., Ribeiro, R.: Utility-based regression. In: Knowledge Discovery in
  Databases: PKDD 2007. pp. 597--604. Springer Berlin Heidelberg (2007)

\bibitem{linex}
Varian, H.R.: A bayesian approach to real estate assessment. Studies in
  Bayesian econometric and statistics in Honor of Leonard J. Savage pp.
  195--208 (1975)

\bibitem{ddl}
Yang, Y., Zha, K., Chen, Y., Wang, H., Katabi, D.: Delving into deep imbalanced
  regression. CoRR  \textbf{abs/2102.09554} (2021),
  \url{https://arxiv.org/abs/2102.09554}

\end{thebibliography}

\newpage 
\appendix

\section{$SERA$ numerical approximation}\label{appendix:sera_app}

$SERA$ and its derivatives 
are approximated by the trapezoidal rule with a uniform grid of $T$ equally spaced intervals with a step of 0.001, as follows.
%
%
%
%
\begin{equation}
\label{eq:seraapprox}
\begin{split}
 SERA &= \int_0^1 SER_{t}\, dt \\
 &\approx \frac{1}{2T}  \sum_{k=1}^T \left(SER_{t_{k-1}} + SER_{t_{k}}\right) \\
     &= \frac{1}{2T}  \left(SER_{t_{0}} + 2SER_{t_{1}} 
     +...+ 2SER_{t_{T-1}} + SER_{t_{T}}\right)  \\
     &= \frac{1}{T}  \bigg(\sum_{k=1}^{T-1} SER_{t_{k}} + \frac{SER_{t_{0}} + SER_{t_{T}}}{2}\bigg)  \\
     &= \frac{1}{T} \bigg(\frac{1}{2}\!\sum_{y_i \in \mathcal{D}^{t_0}}(\hat{y}_{i} - y_i)^2 +\!\sum_{y_i \in \mathcal{D}^{t_1}}(\hat{y}_{i} - y_i)^2 + ... \\
     & \qquad +\!\sum_{y_i \in \mathcal{D}^{t_{T-1}}}(\hat{y}_{i} - y_i)^2 + \frac{1}{2}\!\sum_{y_i \in \mathcal{D}^{t_T}}(\hat{y}_{i} - y_i)^2 \bigg)   
\end{split}
\end{equation}
Similarly, the derivative of $SERA$ w.r.t. a given prediction $\yh_j$ is obtained by
\begin{equation}
\label{eq:seraderiv}
\begin{split}
   \frac{\partial SERA}{\partial \yh_j} &\approx \frac{1}{T}\frac{\partial}{\partial \hat{y}_j} \bigg(\frac{1}{2}\!\sum_{y_i \in \mathcal{D}^{t_0}}(\hat{y}_{i} - y_i)^2 +\!\sum_{y_i \in \mathcal{D}^{t_1}}(\hat{y}_{i} - y_i)^2 + ...\\ & \qquad +\!\sum_{y_i \in \mathcal{D}^{t_{T-1}}}(\hat{y}_{i} - y_i)^2 + \frac{1}{2}\!\sum_{y_i \in \mathcal{D}^{t_T}}(\hat{y}_{i} - y_i)^2 \bigg)\\
    &=\frac{1}{T} \bigg(\sum_{y_i \in \mathcal{D}^{t_0}} (\hat{y}_i - y_i)\delta_{ij} + 2\sum_{y_i \in \mathcal{D}^{t_1}} (\hat{y}_i - y_i)\delta_{ij} + ... \\ 
    & \qquad +\!\sum_{y_i \in \mathcal{D}^{t_{T-1}}} (\hat{y}_i - y_i)\delta_{ij} + \frac{1}{2}\!\sum_{y_i \in \mathcal{D}^{t_T}} (\hat{y}_i - y_i)\delta_{ij}\bigg) \\
     &= \frac{1}{T}  \bigg((\hat{y}_j - y_j) \bigg|_{y_j \in \md^{t_0}} + 2(\hat{y}_j - y_j) \bigg|_{y_j \in \md^{t_1}} + ... \\
   & \qquad + 2(\hat{y}_j - y_j) \bigg|_{y_j \in \md^{t_{T-1}}} + (\hat{y}_j - y_j) \bigg|_{y_j \in \md^{t_T}} \bigg)  \\
    &= \frac{1}{T} \bigg((\yh_j - y_j) \bigg|_{y_j \in \md^{t_0}} + 2\sum_{k=1}^{T-1} (\yh_j - y_j) \bigg|_{y_j \in \md^{t_k}} + (\yh_j - y_j) \bigg|_{y_j \in \md^{t_T}} \bigg)  \\
    &= \frac{1}{T} \bigg(\boldsymbol{1}\left(y_j \in \md^{t_0}\right) + 2\sum_{k=1}^{T-1} \boldsymbol{1}\left(y_j \in \md^{t_k}\right) +  \boldsymbol{1}\left(y_j \in \md^{t_T} \right) \bigg) (\yh_j - y_j)
\end{split}
\end{equation}

\noindent Note that any given instance $y_j$ will always have at least zero relevance, i.e. $\phi(y_j) \geq 0$, so the first term of Equation \eqref{eq:seraderiv} will always be taken into account. Nevertheless, not all the summation terms will be considered for cases where $\phi(y_j) < 1$. With this in mind, we define

\beq n_j = \sum_{k=1}^{K_j} \boldsymbol{1}\left(y_j \in \md^{t_k}\right)~~\eeq

\noindent where $n_j$ is the number of times 
the instance $y_j$ contributes to $SERA$ derivative, where $K_j \in \left[1, T-1\right]$. Equation \eqref{eq:seraderiv} becomes then

\beq \label{eq:sera_deriv_final} \frac{\partial SERA}{\partial \yh_j} \approx \frac{1}{T} \left(1 + 2 n_j +  \boldsymbol{1}\left(y_j \in \md^{t_T} \right) \right) (\yh_j - y_j)~~\eeq

%
%

In this context, the second derivative for a given prediction $\yh_j$ is obtained by

\beq \label{eq:sera_dderiv_final} \frac{\partial^2 SERA}{\partial \yh_j^2} \approx \frac{1}{T} \left(1 + 2 n_j +  \boldsymbol{1}\left(y_j \in \md^{t_T} \right) \right)~~\eeq

We now proceed to a study on the degree of error committed by using the approximations above (Eqs. \eqref{eq:sera_deriv_final}, \eqref{eq:sera_dderiv_final}) against the use of the trapezoidal rule directly on Eqs. \eqref{eq:sera_deriv_trap} and \eqref{eq:sera_dderiv_trap}. For that, we resort to the predictions obtained for \xgbs. The rationale is the following: 1) for each data set, we compute the first and second derivatives using both methods; 2) we calculate the absolute difference between the results obtained from both methods; 3) we average these differences over all instances. 

\begin{figure}
    \centering
    \includegraphics[width=\textwidth]{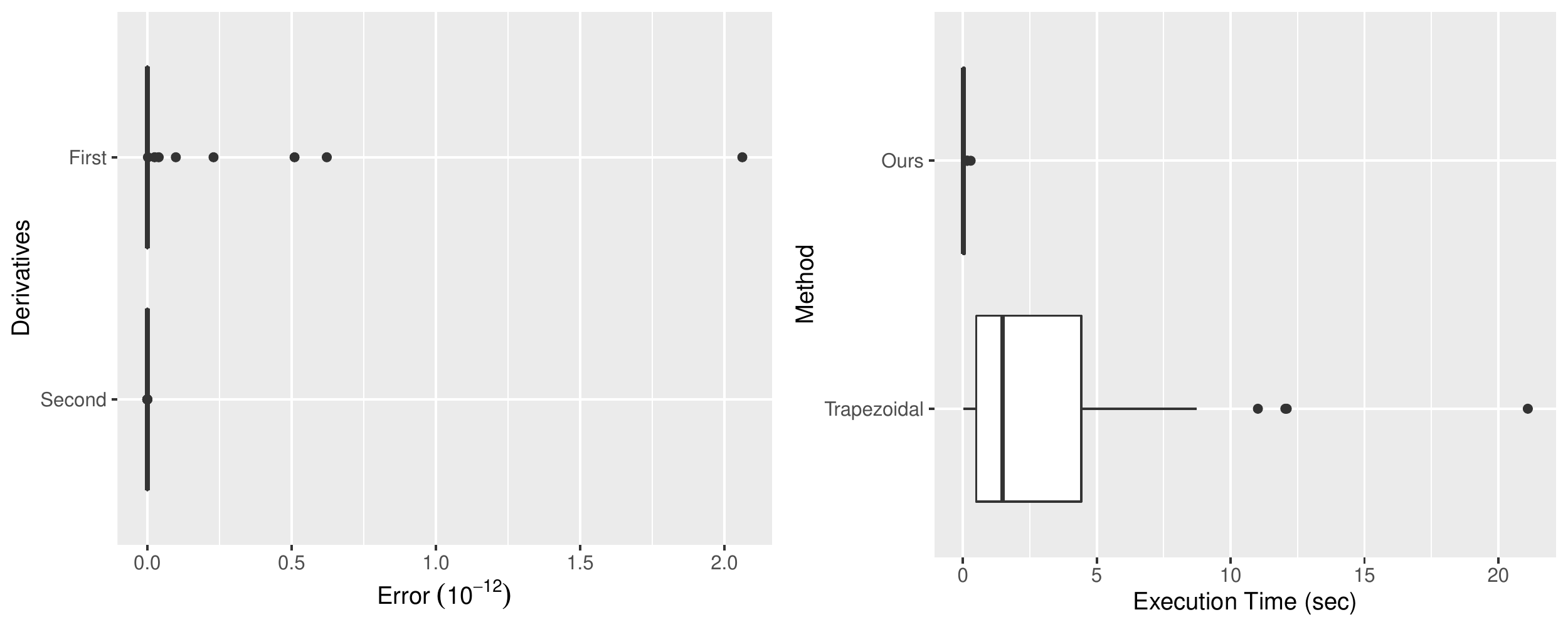}
    \caption{\textbf{Left}: Absolute error differences for the first and second derivatives between our approximations and the trapezoidal rule. \textbf{Right}: Execution time (in seconds) of the first and second derivative for a given data set, under our approximation and the trapezoidal rule.}
    \label{fig:approxs}
\end{figure}

The results obtained using this evaluation are depicted in the left box plot of Figure \ref{fig:approxs}.

Results show that both first and second derivative approximations have a minor error ($\approx 10^{-12}$). 
Sim On the right box plot of Figure \ref{fig:approxs}, we also show the difference in execution time using both methods for each data set. Here, the execution time is measured as the time taken to evaluate both the first and second derivatives. As we can see, there is a non-negligible difference between our approximation and the trapezoidal rule as the size of a data set increases. 

\section{Tables of Results}\label{appendix:results}

In this section, we report the $MSE$ and $SERA$ results obtained in out-of-sample of each data set.  

\begin{table}[!hbt]
\caption{$MSE$ results in out-of-sample, with the best models per data set in bold.
Model (\#wins): \xgbm~(27), \xgbs~(2), \lgbmm~(6), \lgbms~(1).}
\label{tab:results2}
\scriptsize
\begin{minipage}{0.49\linewidth}
\resizebox{\textwidth}{!}{
\begin{tabular}{ccccc}
\toprule
id & \xgbs & \xgbm & \lgbms & \lgbmm \\ 
\midrule
1 & 6.41e-01 & \textbf{3.36e-01} & 5.56e-01 & 5.15e-01 \\ 
2 & \textbf{1.00e-02} & 1.41e-02 & 1.42e-02 & 1.55e-02 \\ 
3 & 4.60e+01 & 4.76e+01 & 4.67e+01 & \textbf{4.56e+01} \\ 
4 & 8.64e+07 & 2.39e+07 & 6.94e+07 & \textbf{1.60e+07} \\ 
5 & 9.81e+05 & \textbf{3.34e+05} & 9.18e+05 & 3.75e+05 \\ 
6 & 1.44e+01 & \textbf{1.08e+01} & 2.25e+01 & 1.84e+01 \\ 
7 & 1.68e+04 & 4.62e+03 & \textbf{1.82e+02} & 4.29e+02 \\ 
8 & 2.30e+01 & 1.78e+01 & 2.36e+01 & \textbf{1.69e+01} \\ 
9 & 1.23e+06 & 4.88e+05 & 1.32e+06 & \textbf{4.80e+05} \\ 
10 & 1.98e+00 & \textbf{2.57e-02} & 6.84e-01 & 7.49e-01 \\ 
11 & 3.42e+00 & \textbf{8.19e-02} & 1.28e+00 & 9.15e-01 \\ 
12 & 3.46e-01 & \textbf{3.00e-01} & 3.24e-01 & 3.10e-01 \\ 
13 & 1.20e+01 & \textbf{2.17e+00} & 7.59e+01 & 1.65e+02 \\ 
14 & 7.72e-01 & \textbf{4.87e-01} & 2.13e+00 & 2.14e+00 \\ 
15 & 2.37e-01 & \textbf{1.11e-01} & 8.85e-01 & 7.62e-01 \\ 
16 & 5.27e+01 & \textbf{2.53e+01} & 2.23e+02 & 2.53e+02 \\ 
17 & 4.21e+01 & \textbf{1.33e+01} & 7.23e+02 & 1.02e+03 \\ 
18 & 2.24e-02 & \textbf{4.66e-03} & 2.16e-02 & 8.45e-03 \\ 
\bottomrule
\end{tabular}
}
\end{minipage} 
\begin{minipage}{0.47\linewidth}
\resizebox{\textwidth}{!}{
\begin{tabular}{ccccc}
\toprule
id & \xgbs & \xgbm & \lgbms & \lgbmm \\ 
\midrule
19 & 1.46e-02 & \textbf{9.74e-03} & 1.59e-02 & 1.41e-02 \\ 
20 & 2.93e+00 & \textbf{2.35e+00} & 2.72e+00 & 2.45e+00 \\ 
21 & 6.66e+00 & \textbf{4.87e+00} & 5.96e+00 & 5.59e+00 \\ 
22 & 2.02e+00 & \textbf{3.48e-01} & 9.53e-01 & 7.85e-01 \\ 
23 & 4.78e-08 & 2.67e-08 & 3.36e-08 & \textbf{2.47e-08} \\ 
24 & 2.06e+00 & \textbf{6.77e-01} & 4.43e+01 & 4.18e+01 \\ 
25 & 2.00e+01 & \textbf{4.52e+00} & 1.25e+02 & 8.32e+01 \\ 
26 & 2.36e-03 & \textbf{2.00e-03} & 1.01e-02 & 1.06e-02 \\ 
27 & 7.12e-02 & \textbf{6.56e-02} & 1.00e-01 & 1.13e-01 \\ 
28 & 8.44e-05 & \textbf{6.40e-05} & 8.02e-05 & 6.76e-05 \\ 
29 & 7.93e-06 & 2.01e-06 & 9.01e-06 & \textbf{1.99e-06} \\ 
30 & 2.07e-03 & \textbf{1.93e-03} & 2.64e-03 & 2.89e-03 \\ 
31 & 3.89e+02 & \textbf{2.16e+02} & 1.37e+03 & 1.23e+03 \\ 
32 & 5.85e-06 & \textbf{4.37e-06} & 1.65e-05 & 1.39e-05 \\ 
33 & 8.39e+09 & \textbf{3.38e+09} & 1.19e+10 & 1.47e+10 \\ 
34 & 1.49e+09 & \textbf{1.09e+09} & 1.38e+09 & 1.13e+09 \\ 
35 & 1.23e+09 & \textbf{8.63e+08} & 1.12e+09 & 9.48e+08 \\ 
36 & \textbf{9.89e+07} & 1.01e+08 & 1.07e+08 & 9.95e+07 \\ 
\bottomrule
\end{tabular}
}
\end{minipage}
\end{table}

\begin{table}[t]
\caption{$SERA$ results in out-of-sample, with the best models per data set in bold.
Model (\#wins): \xgbm~(16), \xgbs~(14), \lgbms~(5), \lgbmm~(1).}
\label{tab:results1}
\scriptsize
\begin{minipage}{0.49\linewidth}
\resizebox{\textwidth}{!}{
\begin{tabular}{ccccc}
\toprule
id & \xgbs & \xgbm & \lgbms & \lgbmm \\ 
\midrule
1 & 3.86e+00 & \textbf{1.48e+00} & 3.36e+00 & 2.13e+00 \\ 
2 & \textbf{1.47e-01} & 2.50e-01 & 2.03e-01 & 2.66e-01 \\ 
3 & 1.70e+03 & 1.75e+03 & \textbf{1.58e+03} & 1.61e+03 \\ 
4 & 1.18e+09 & 3.63e+08 & 9.99e+08 & \textbf{1.87e+08} \\ 
5 & 3.50e+07 & \textbf{1.34e+07} & 3.59e+07 & 1.76e+07 \\ 
6 & \textbf{6.39e+02} & 6.47e+02 & 9.58e+02 & 1.15e+03 \\ 
7 & 5.82e+05 & 2.27e+05 & \textbf{4.66e+03} & 1.41e+04 \\ 
8 & \textbf{7.06e+02} & 9.11e+02 & 9.06e+02 & 8.94e+02 \\ 
9 & 8.70e+07 &  \textbf{5.47e+07} & 8.47e+07 & 5.48e+07 \\ 
10 & 1.90e+02 & \textbf{1.91e+00} & 4.43e+01 & 4.86e+01 \\ 
11 & 4.22e+02 & \textbf{1.12e+01} & 1.32e+02 & 8.08e+01 \\ 
12 & 3.03e+01 & 3.24e+01 & \textbf{2.32e+01} & 3.57e+01 \\ 
13 & 1.07e+03 & \textbf{2.47e+02} & 1.05e+04 & 1.59e+04 \\ 
14 & 5.11e+01 & \textbf{4.35e+01} & 2.03e+02 & 2.34e+02 \\ 
15 & 1.63e+01 & \textbf{1.46e+01} & 9.51e+01 & 9.17e+01 \\ 
16 & \textbf{5.77e+03} & 6.72e+03 & 4.26e+04 & 5.02e+04 \\ 
17 & 3.52e+03 & 3.09e+03 & \textbf{1.29e+05} & 2.12e+05 \\ 
18 & 4.71e+00 & 1.14e+00 & \textbf{4.94e+00} & 2.59e+00 \\ 
\bottomrule
\end{tabular}
}
\end{minipage} 
\begin{minipage}{0.47\linewidth}
\resizebox{\textwidth}{!}{
\begin{tabular}{ccccc}
\toprule
id & \xgbs & \xgbm & \lgbms & \lgbmm \\ 
\midrule
19 & 2.38e+00 & \textbf{2.30e+00} & 2.90e+00 & 3.58e+00 \\ 
20 & \textbf{4.84e+02} & 6.05e+02 & 5.47e+02 & 7.81e+02 \\ 
21 & \textbf{1.92e+03} & 2.26e+03 & 2.25e+03 & 2.86e+03 \\ 
22 & \textbf{4.30e+01} & 1.54e+02 & 1.70e+02 & 5.50e+02 \\ 
23 & 1.14e-05 & 1.65e-05 & \textbf{1.11e-05} & 1.62e-05 \\ 
24 & 4.05e+02 & \textbf{3.48e+02} & 2.25e+04 & 2.63e+04 \\ 
25 & 1.01e+04 & \textbf{1.58e+03} & 5.71e+04 & 4.39e+04 \\ 
26 & \textbf{9.31e-01} & 1.12e+00 & 4.27e+00 & 5.02e+00 \\ 
27 & \textbf{2.90e+01} & 3.82e+01 & 5.48e+01 & 7.93e+01 \\ 
28 & \textbf{2.24e-02} & 2.82e-02 & 3.13e-02 & 4.05e-02 \\ 
29 & 1.78e-03 & \textbf{1.33e-03} & 2.04e-03 & 1.47e-03 \\ 
30 & \textbf{2.68e+00} & 2.93e+00 & 3.66e+00 & 4.61e+00 \\ 
31 & 2.77e+05 & \textbf{2.23e+05} & 1.93e+06 & 2.21e+06 \\ 
32 & \textbf{5.84e-03} & 6.20e-03 & 1.87e-02 & 1.98e-02 \\ 
33 & 1.21e+13 & \textbf{5.64e+12} & 1.61e+13 & 2.36e+13 \\ 
34 & \textbf{2.82e+12} & 3.25e+12 & 3.08e+12 & 3.85e+12 \\ 
35 & \textbf{2.24e+12} & 2.33e+12 & 2.30e+12 & 3.07e+12 \\ 
36 & 7.57e+11 & 7.81e+11 & \textbf{7.24e+11} & 7.46e+11 \\
\bottomrule
\end{tabular}
}
\end{minipage}
\end{table}
\end{document}